\documentclass[review]{elsarticle}

\usepackage{graphicx}
\usepackage{lineno,hyperref}
\usepackage{caption}
\usepackage{subcaption}
\usepackage{amsmath}
\usepackage{multirow}
\usepackage{float}
\usepackage[table,xcdraw]{xcolor}
\modulolinenumbers[5]

\DeclareMathOperator*{\argmax}{arg\,max} 

\journal{Journal of \LaTeX\ Templates}

\begin{document}

\begin{frontmatter}

\title{Automatic Identification of \textit{Scenedesmus} Polymorphic Microalgae from Microscopic Images}

\author[affiliation1]{Jhony-Heriberto Giraldo-Zuluaga \corref{mycorrespondingauthor}}
\cortext[mycorrespondingauthor]{Corresponding author}
\ead{jhonygiraldoz@gmail.com, heriberto.giraldo@udea.edu.co}

\author[affiliation1]{Geman Diez}

\author[affiliation1]{Alexander G\'omez}

\author[affiliation2]{Tatiana Mart\'inez}

\author[affiliation2]{Mariana Pe\~nuela V\'asquez}

\author[affiliation1]{Jes\'us Francisco Vargas Bonilla}

\author[affiliation1]{Augusto Enrique Salazar Jim\'enez}

\address[affiliation1]{Grupo de Investigaci\'on SISTEMIC, Facultad de Ingenier\'ia, Universidad de Antioquia UdeA, Calle 70 No. 52-21, Medell\'in, Colombia}
\address[affiliation2]{Grupo de Investigaci\'on de Bioprocesos, Facultad de Ingenier\'ia, Universidad de Antioquia UdeA, Calle 70 No. 52-21, Medell\'in, Colombia}

\begin{abstract}

Microalgae counting is used to measure biomass quantity. Usually, it is performed in a manual way using a Neubauer chamber and expert criterion, with the risk of a high error rate. This paper addresses the methodology for automatic identification of Scenedesmus microalgae (used in the methane production and food industry) and applies it to images captured by a digital microscope. The use of contrast adaptive histogram equalization for pre-processing, and active contours for segmentation are presented. The calculation of statistical features (Histogram of Oriented Gradients, Hu and Zernike moments) with texture features (Haralick and Local Binary Patterns descriptors) are proposed for algae characterization. Scenedesmus algae can build coenobia consisting of 1, 2, 4 and 8 cells. The amount of algae of each coenobium helps to determine the amount of lipids, proteins, and other substances in a given sample of a algae crop. The knowledge of the quantity of those elements improves the quality of bioprocess applications. Classification of coenobia achieves accuracies of 98.63\% and 97.32\% with Support Vector Machine (SVM) and Artificial Neural Network (ANN), respectively. According to the results it is possible to consider the proposed methodology as an alternative to the traditional technique for algae counting. The database used in this paper is publicly available for download.

\end{abstract}


\end{frontmatter}


\section{Introduction}

\textit{Scenedesmus} microalgae are used in the methane \cite{alonso1995microalgas} and pigment production. In the food industry they are used to produce antioxidants whose high consumption is intended to reduce the risk of cancer. \textit{Scenedesmus} also has been used in biodiesel production from lipid synthesis \cite{wu2013enhancement}, and absorption of heavy metals from Blackwater or from other habitats with minimal nutrients \cite{terry2002biosorption,gorbi2006differential,hodaifa2013biomass}. \textit{Scenedesmus} microalgae has a biotechnological interest by its morphological capacity for adapting to environment changes. \textit{Scenedesmus} can change their cellular organization and form coenobia consisting of 1, 2, 4, 8 and 16 cells. This phenomenon is known as polymorphism and depends on and is caused by environmental and stress-inducing conditions.

A common procedure in microalgae biomass research includes a counting process. Through this counting process it is possible to determine the level of lipids, carotenoids, proteins and other substances of interest in a given sample. The manual counting method with a microscope and a Neubauer hemocytometer is commonly used to determine those important levels. Visual features like shape, size and color, but also the image context help to identify the microorganisms present in the image.

In the manual counting process, the microalgae are counted and classified according to the amount of cells per coenobium. These could be 1, 2, 4, 8 or 16 cells due to the growth conditions. Each square of the Neubauer chamber is tabulated and the expert counts microalgae per group, based on his own criterion. Since this is a method that requires human supervision, the results can suffer an error between 10\% and 22\% \cite{cartwright1973laboratorio}. An automatic counting procedure could optimize the experts' time letting them concentrating  on data only.

The most used automated method is the electric impedance, which is based on the changes of the electric resistors produced by cells suspended in a liquid conductor. This method has a known limitation when cells are overlapped, because it is not possible to record independent cells but the amount of biomass present in the sample. This method is usually not used in biotechnology, because the counting searches to register biomass incrementation \cite{maya2007hemograma,arredondo2007metodos}, but does not give detailed information about the type of cells in the sample. The electric method helps to identify the amount of biological mass in the sample, but not the type of cells or cell conglomerates, hence does not let to identify the amount of 1, 2, 4, 8, 16 cell coenobia in the sample.


Digital image processing and artificial vision techniques offer a broad number of solutions to the biology field including the microscopy field and cell analysis. In computer vision the classification of microorganisms can be separated in finding each organism of interest on the image (segmentation), extracting an characteristic pattern (feature extraction), and classifying them in a set of predefined classes (classification). An image, like any signal, can be corrupted by noise or not be in ideal conditions, hence it is common to use a pre-processing stage before any other step to enhance classification.

In general microscope-captured images have no uniform illumination, state of focus, overlapping, and similarity with the background. Those issues can influence the automatic segmentation quality \cite{santhi2013automatic,gupta2012image}. Histogram Equalization is commonly used to improve low contrast images \cite{gonzalez2004digital} and morphological filters highlight tiny image elements \cite{santhi2013automatic}. Once the interest components are underlined, threshold techniques based on Laplace of the Gaussian (LOG) operator, Canny Edge Detector algorithm or any high-pass filter are common approaches to segment regions of interest.

Features to classify microorganisms are related to shape and frequency features, different shape descriptors have been used to characterize microorganism forms, major and minor axis ratio, inner area and shape perimeter provides an idea of the object characteristics \cite{mosleh2012preliminary}, Fourier features, Hu and Zernike moments are broadly used for this task \cite{santhi2013automatic,thiel1996automated}. Color and texture are also used if the studied objects have inner structures that the expert uses to recognize them \cite{mansoor2011automatic}.

In microalgae several application on classifying species with computer vision has been done. Santhi et al. achieved 98\% of accuracy classifying five types of algae: Diatom, Closterium, acerosum, Pediastrum and Pinnularia with object size, shape, and texture based on feature extraction techniques \cite{santhi2013automatic}. Luo et al. use circular shape diatoms to identify using canny filter, Fourier spectrum descriptors and Artificial Neural Networks with 94.44\% of accuracy \cite{luo2011automatic}. Fluorescence in microalgae was employed by Walker et al. to segment using region growing, and classifying them taking advantage of algae shape, frequency domain and second order statistical properties \cite{walker2002fluorescence}. Molesh et al. got 93\% of accuracy classifying five types of algae: Navicula Scenedesmus Microcystis Oscillatoria and Chroococcus in river water also using shape and texture descriptors \cite{mosleh2012preliminary}.

This work, unlike other approaches, confront a classification inside the same species of microalgae. We propose an automatic classification system for \textit{Scene-desmus} Polymorphic Microalgae of 1, 2, 4, 8 coenobia. Our approach uses a contrast correction technique and active contours, and is an energy minimization segmentation method. For classification Hu, Zernicke Histogram of Oriented Gradients (HOG), Local Binary Pattern (LBP) and Haralick descriptors fed a support vector machine and an Artificial Neural Network. The database used in this paper is publicly available for download from the web page of the project \cite{Salazar2015microalgaeDatabase}

Material and methods are mentioned in Section 2. Section 3 describes the experimental framework used to test the model. Section 4 discusses the results. Finally, in Section 5 conclusions and future work are presented.

\section{Methods}

\begin{figure}[h]
\centering
\includegraphics[width=0.8\textwidth]{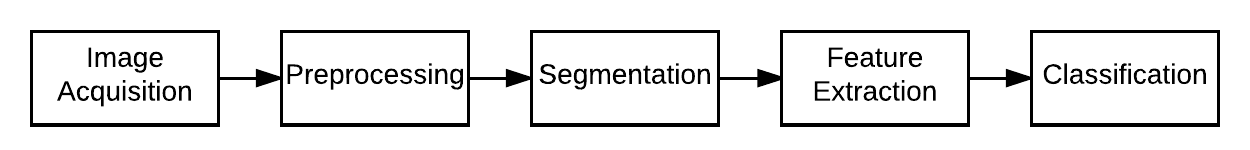}
\caption{Methodology.}
\label{fig:methodology}
\end{figure}

Here, the proposed method is described, as shown in Figure \ref{fig:methodology}, where the method is subdivided in its main processes. The pre-processing step highlights characteristics and helps to enhance the segmentation stage. The segmentation found individual Scenedesmus algae. The last two procedures classify each segmented alga as 1, 2, 4 or 8 coenobium.

\subsection{Image Acquisition}

\begin{figure}[h]
    \centering
    \begin{subfigure}[b]{0.45\textwidth}
		\includegraphics[width=\textwidth]{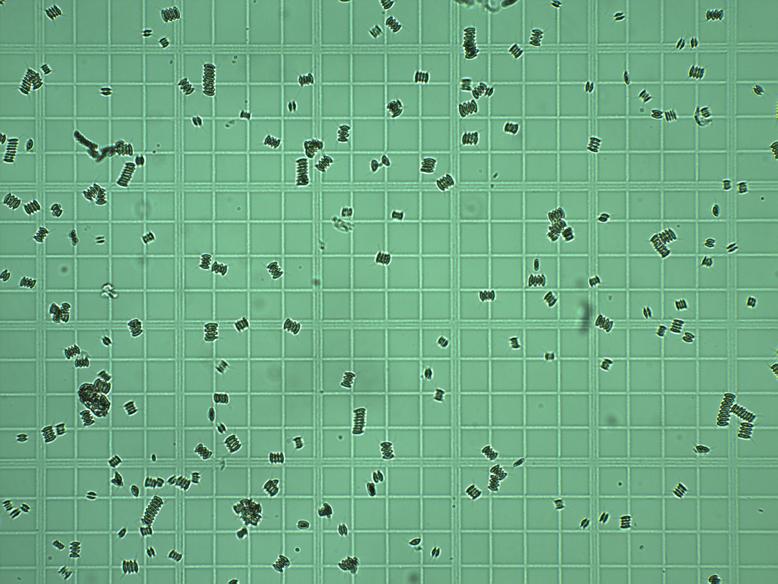}
        \caption{}
        \label{fig:neubauerSuperior}
    \end{subfigure}
    \begin{subfigure}[b]{0.45\textwidth}
		\includegraphics[width=\textwidth]{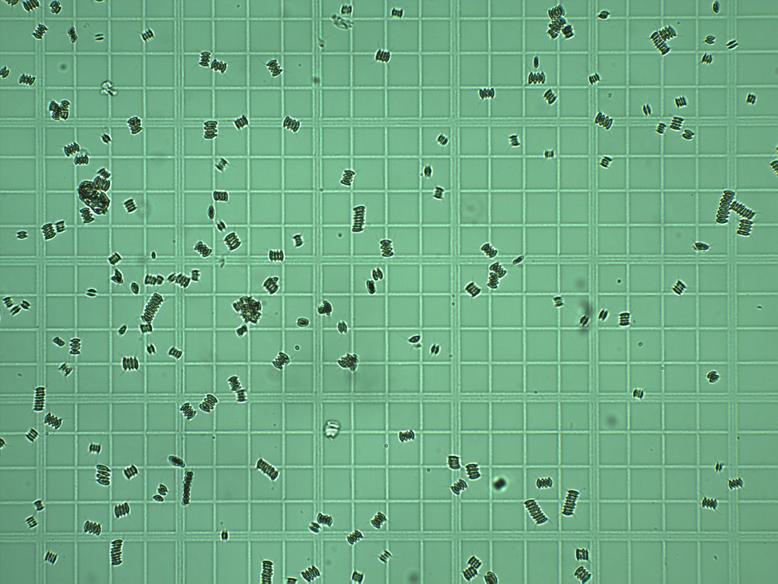}
        \caption{}
        \label{fig:neubauerInferior}
    \end{subfigure}
    \caption{Microscopic images. (a) Superior image. (b) Inferior image.}
    \label{fig:neubauer}
\end{figure}

The sample images of \textit{Scenedesmus sp} were obtained from the algae bank of Laboratorio de Bioprocesos of Universidad de Antioquia. Those samples were donated by Universidad de Zulia in Venezuela \cite{quevedo2008crecimiento}. The samples are frozen in Eppendorf tubes after the cultivation process in order to ensure the conservation of the algae. When the sample images are taken, it is necessary to agitate the sample to obtain a uniform distribution of the algae, then a sample of 10 milliliters from each tube is put in a Neubauer chamber. The region of interest is the 25 central squares of the chamber. Due to hardware limitations in the image acquisition it is necessary take two images to cover the whole study area. Figure \ref{fig:neubauer} shows those two images. The counting process is performed taking into account the zone where this two images concur and standard counting rules that link position and number of algae in a image region \cite{vega2007metodos}.

\subsection{Preprocess}

Microscope images offer typically a set of conditions that make the identification and counting task more difficult. Conditions like blur, non-homogeneity of the background and salt-pepper noise, affect the performance of classification and counting algorithms. Other image elements like the grid of the Neubauer chamber provides important information to check the counting rules on the manual counting process. In order to correct these image acquisition effects the following methods are proposed.

\subsubsection{Histogram Equalization}

\begin{figure}[h]
    \centering
    \begin{subfigure}[b]{0.4\textwidth}
		\includegraphics[width=\textwidth]{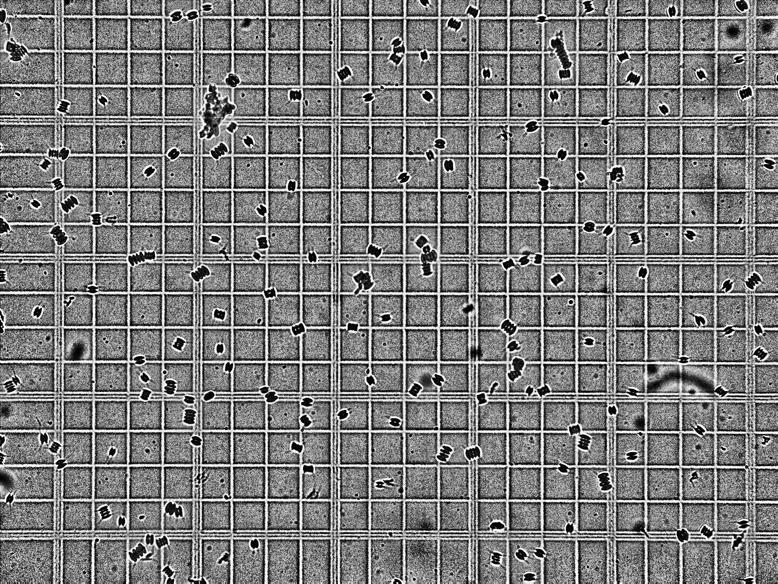}
        \caption{}
        \label{fig:afterClahe}
    \end{subfigure}
    \begin{subfigure}[b]{0.4\textwidth}
		\includegraphics[width=\textwidth]{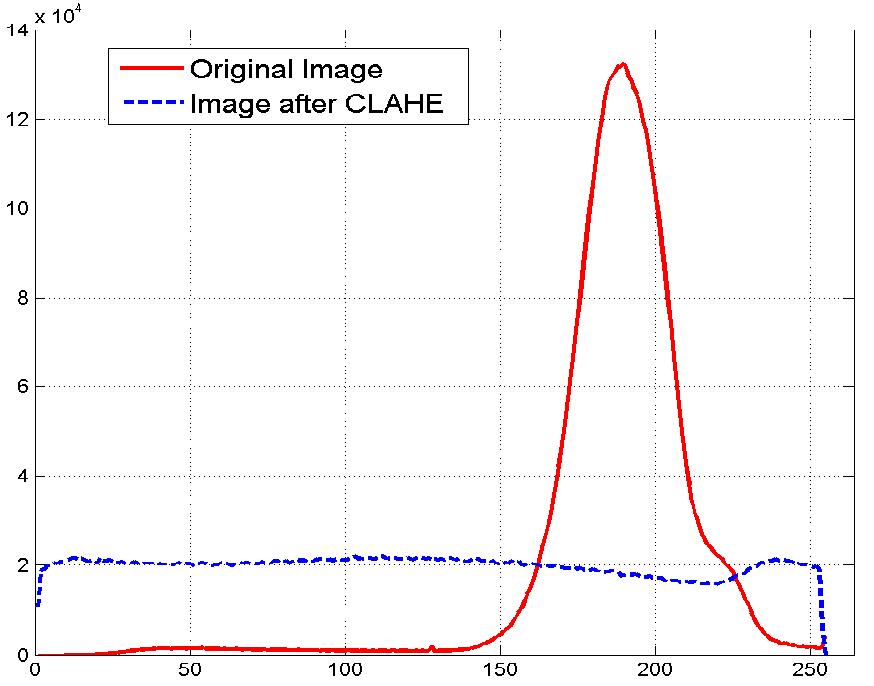}
        \caption{}
        \label{fig:histImage}
    \end{subfigure}
    \caption{CLAHE Results. (a) Output CLAHE image. (b) Histrogram of the raw and output image.}
    \label{fig:clahe}
\end{figure}

Light inhomogeneity affects the reconstruction of the Neubauer chamber lines. Histogram equalization is a common used method to improve these image conditions. In this work Contrast Limited Adaptive Histogram Equalization (CLAHE) was used as an image equalization method. CLAHE finds the mapping for each pixel based on the intensity values surrounding pixel on a local neighborhood \cite{thiel1995automated}. Figure \ref{fig:afterClahe} shows the outcome obtained by the use of this Histogram equalization method, and Figure \ref{fig:histImage} shows the histogram map of the original and the equalized image.

This histogram equalization step guarantees the grid reconstruction, which is necessary, because of  the difference between the grid, algae, and background, in comparison with the original image.

\subsubsection{Color Quantization}

Color Quantization methods reduce the number of colors with which an image can be represented. This reduction in the quantization levels increases the color distance between algae and background. Let $I_g$ an image with 256 gray levels and $n_{L+1}$ the number of levels used in the color quantization. The Equation \ref{eqn:posterImage} shows the posterized image transformation.

\begin{equation}
I_p(x,y)= round\left(\displaystyle round\left(\displaystyle \frac{I_g(x,y)n_L}{255} \right)\left(\displaystyle \frac{255}{n_L} \right)\right)
\label{eqn:posterImage}
\end{equation}

\begin{figure}[h]
    \centering
    \begin{subfigure}[b]{0.3\textwidth}
		\includegraphics[width=\textwidth]{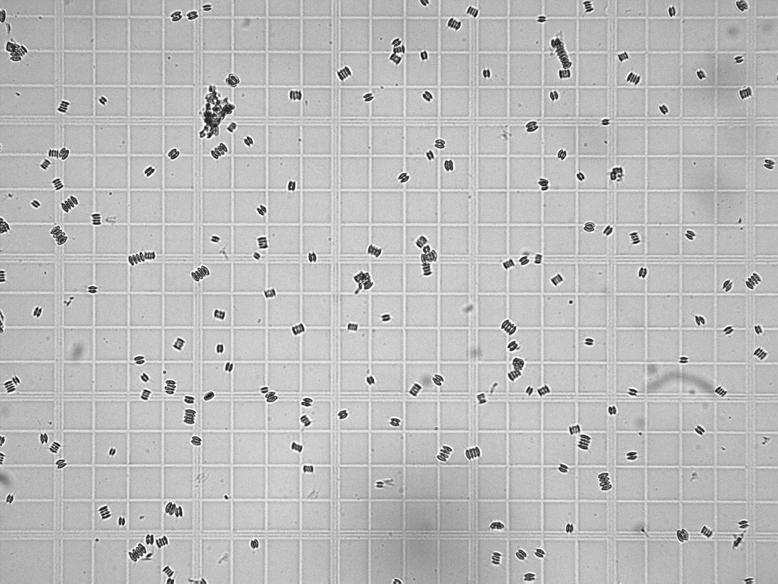}
        \caption{}
        \label{fig:quantOriginal}
    \end{subfigure}
    \begin{subfigure}[b]{0.3\textwidth}
		\includegraphics[width=\textwidth]{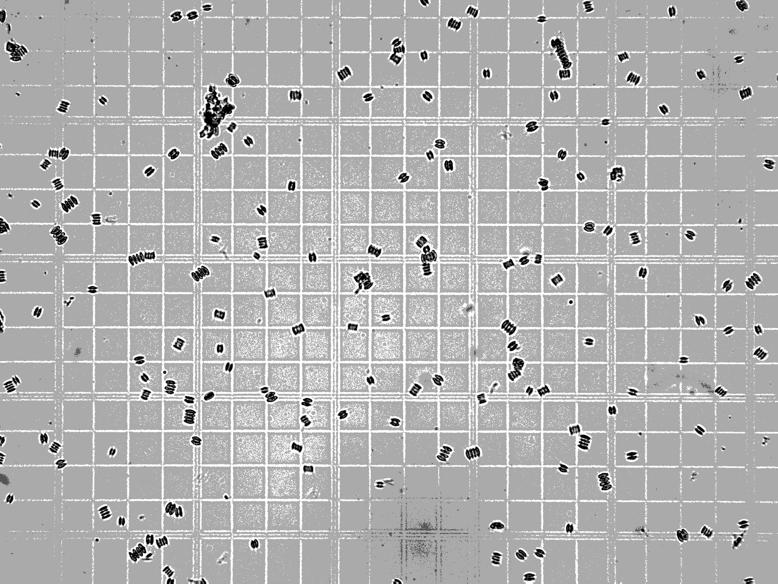}
        \caption{}
        \label{fig:posterImage}
    \end{subfigure}
    \begin{subfigure}[b]{0.3\textwidth}
		\includegraphics[width=\textwidth]{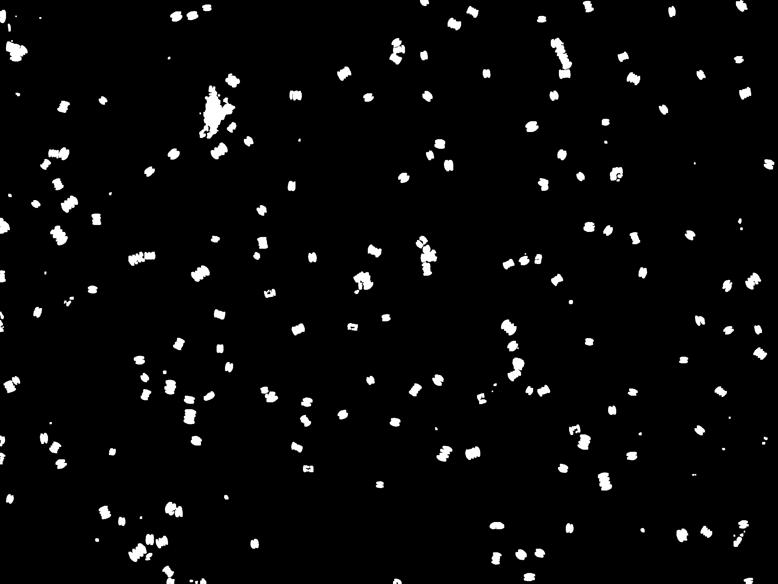}
        \caption{}
        \label{fig:quantMask}
    \end{subfigure}
    \caption{Color Quantization Results. (a) Raw image. (b) Image with saturation correction and color quantization. (c) Otsu binarization.}
    \label{fig:quant}
\end{figure}

In this paper, $n_L = 3$ was chosen. When $n_L < 3$ intensities values of the microalgae are very similar to background intensities. On the other hand, when $n_L > 3$ ,the salt and pepper noise is highlighted. The image posterization using $n_L = 3$ allows to eliminate background image and undesirable lighting effects. Figure \ref{fig:posterImage} shows the result of applying this color quantization on a typical image. The difference between algae and background lets to use an Otsu threshold to construct a mask. Figure \ref{fig:quantMask} shows the Otsu image result. Otsu binarization can be used as a seed for another algorithm to improve the segmentation task.

\subsection{Segmentation}

\begin{figure}[h]
\centering
\includegraphics[width=\textwidth]{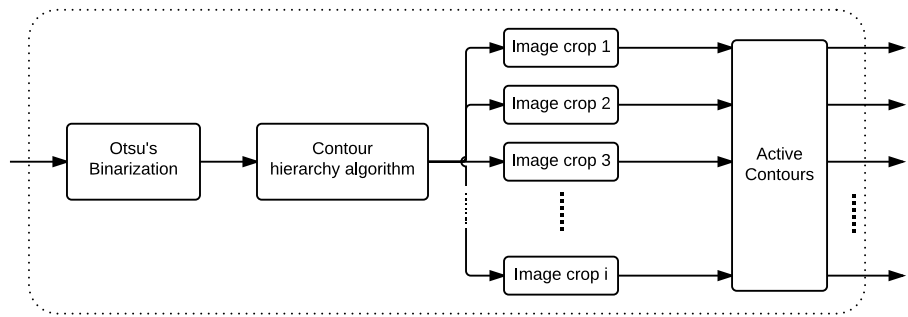}
\caption{Algae segmentation diagram block.}
\label{fig:segDiag}
\end{figure}

In this section the segmentation process is explained. Figure \ref{fig:segDiag} shows the segmentation process. The contour hierarchy block selects images that contain possible algae based on size and the presence of internal contour. These are common in corrupted elements in the sample, those algae are extracted and put in the active contours block.

Color quantization allows the color separation of algae from background, in order to find a proper global threshold value that always discriminating objects from the background, which is a difficult task due to the natural variation of algae under microscope circumstances. An Otsu's binarization was used. The Otsu method assumes binarization like a bi-class clustering problem and selects a threshold value that minimizes intra-class variation. Figure \ref{fig:quantMask} shows the final output of the Otsu method.

\begin{figure}[hbtp]
\centering
\includegraphics[width=0.6\textwidth]{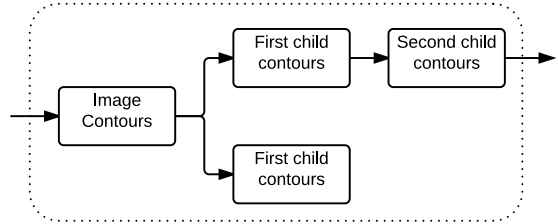}
\caption{Contour hierarchy algorithm.}
\label{fig:contours}
\end{figure}

Microalgae present shapes inside contours usually, for this reason image contours were classified in a parent-child hierarchy. Figure \ref{fig:contours} shows the parent-child hierarchy. Contours whose parent was main image contour and had children contours are candidates of microalgae. 

\begin{figure}[h]
    \centering
    \begin{subfigure}[b]{0.2\textwidth}
		\includegraphics[width=\textwidth]{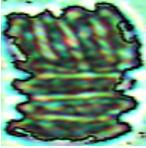}
        \caption{}
        \label{fig:ov1}
    \end{subfigure}
    \begin{subfigure}[b]{0.2\textwidth}
		\includegraphics[width=\textwidth]{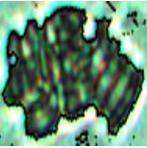}
        \caption{}
        \label{fig:ov2}
    \end{subfigure}
    \begin{subfigure}[b]{0.2\textwidth}
		\includegraphics[width=\textwidth]{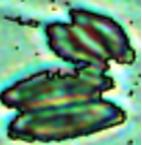}
        \caption{}
        \label{fig:ov3}
    \end{subfigure}
    \begin{subfigure}[b]{0.2\textwidth}
		\includegraphics[width=\textwidth]{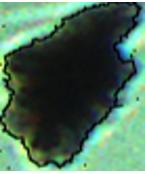}
        \caption{}
        \label{fig:ov4}
    \end{subfigure}
    \caption{(a) Overlapping microalgae. (b) Microalgae pile. (c) Linked microalgae. (d) Unusual shapes.}
    \label{fig:cases}
\end{figure}

Although the contour method can find all microalgae on the image, if the samples had saved considerable time in a refrigerator, they could have been contaminated by other microorganisms, which are in laboratory or in the environment. Noise in the image like bubbles, dead microalgae or other microorganism can be present on the sample. Figure \ref{fig:cases} shows some of these special cases present in the database. This special cases were removed from the database for the classification experiments that will be explained in the Section \ref{sec:experiments}.

Candidate contours have random orientation. Some shape descriptors may vary their values if the input samples do not have the same orientation. The orientation of segmented algae was standardized aligning each contour with respect to its image borders. Sobel filter and Fourier Transform (FT) were applied on each image in order to rotate all images in the same orientation. The rotation property of the 2D FT shows that if the main vertical and horizontal components of an object are rotated in space the shift is reflected on the frequency domain \cite{gonzalez2004digital}. A linear regression was used to estimate the orientation of the spectrum and the angle needed to align the image.

The output images of the automatic alignment procedure based on the Fourier transform do not have the same orientation. A lot of microalgae have a vertical orientation, others horizontal, and others have a random orientation (specially on one coenobium microalgae). The features chosen, like LBP, Haralick, HOG, and Zernike are dependent on the orientation. The extracted features will be explained in the section \ref{featuresExtraction}. Due to the theoretical orientation sensibility on the descriptors, some contours have to be manually aligned in a vertical way in order to test if the classification performance is affected by this rotation as explained in Section \ref{sec:experiments}.

\begin{figure}[h]
    \centering
    \begin{subfigure}[b]{0.3\textwidth}
		\includegraphics[width=\textwidth]{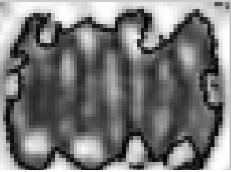}
        \caption{}
        \label{fig:ac1}
    \end{subfigure}
    \begin{subfigure}[b]{0.3\textwidth}
		\includegraphics[width=\textwidth]{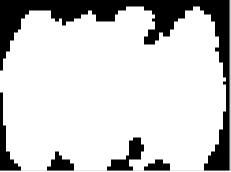}
        \caption{}
        \label{fig:ac2}
    \end{subfigure}
    \caption{Contour image segmentation. (a) Original Image. (b) Segmented Image.}
    \label{fig:ac}
\end{figure}

To extract shapes and features of microalgae, first the borders of the microalgae must be extracted. This task is performed taking into account that microalgae borders are not uniform in color and shape. For that reason active contours are used to segment the algae contours due to this technique adapt a spline function based on an energy-minimization, based on external image forces like edges, gradient, intensities \cite{kass1988snakes}. This energy minimization approach makes  the border segmentation possible regardless the high variability between the algae borders. Figure \ref{fig:ac} shows an example result of the active contour procedure. 

\subsection{Features}\label{featuresExtraction}

In this section  the shape and texture descriptors will be explained. Although it is difficult to give a precise definition of texture, this is an area organization phenomenon present in all pictures. These phenomena are described in therms of pixel bright intensities and organization patterns. Algae usually share common elements like internal structures that help experts to identify the different types of cells, for that reason texture analysis can be considered to ease this classification problem.

\subsubsection{Local Binary Patterns}

The Local Binary Patterns or LBP texture descriptors are used in computer vision as a classification feature \cite{wang1990texture,he1990texture}. LBP describes the texture of the image taking into account the neighborhood of each pixel. LBP is a popular technique in computer vision due to its discriminative power and computational simplicity that allow us to apply them in our application because of the large number of algae present on each image.

\subsubsection{Haralick descriptors}

The Haralick descriptors are 14 features extracted from the co-occurrence matrix ($\vec{P}$). The co-occurrence matrix is a square matrix $\vec{P} \in R^{gxg}$, where $g$ is the numbers of gray levels in the image. This matrix $\vec{P}$ considers the probability of a pixel with value i be adjacent to a pixel with value j. The 14 textural features were described for Haralick et al. on \cite{haralick1973textural}.

\subsubsection{Histogram of oriented gradients}

The histogram of oriented gradients(HOG) was proposed by Dalal and Triggs \cite{dalal2005histograms}. The main idea is that local object shape could be represented by the distribution of local intensity gradients or edges directions, even without precise knowledge about the corresponding gradient or edge positions. To compute the HOG features, the gradient magnitude is calculated and orientation values from brightness L of each pixel.

\subsubsection{Hu moments}

The invariant moments was proposed by Hu \cite{hu1962visual}, the basis idea is obtain invariant descriptors to rotation, translation and scale. Like the HOG features, the Hu moments are useful in the pattern recognition field. There are seven Hu moments.

\subsubsection{Zernike moments}

The Zernike polynomials were proposed by Zernike \cite{von1934beugungstheorie}. They were used to represent the optical aberration, but found application in pattern recognition \cite{teague1980image}. The Zernike moments are mathematical descriptors with some mathematical properties. They have rotational invariant properties, but normalizing mass center and scale the radius, those moments can be scale and translation invariant \cite{khotanzad1990invariant}.

\subsection{Feature Selection}

On this section the used feature selection algorithm  will be explained. The feature selection process is an important step in model construction, where redundant data is removed without loss of information. The feature selection simplifies the model, makes the training times shorter, enhance generalization by reducing over fitting. We used the Sequential Forward Selection (SFS) in this paper as feature selector.


\subsubsection{Sequential Forward Selection}

The SFS is the simplest greedy search algorithm for feature selection. Starting from the empty set, sequentially add the feature $x^+$ that maximize $J(Y_k+x^+)$ when combined with the features $Y_k$ that have already been selected, where the $J$ function is the classification method chosen. On each iteration the SFS obtain the best feature based on some supervised classifier with respect to a database \cite{ruckstiess2011sequential}.

\begin{equation}
x^+ = \argmax_{x \notin Y_k}(J(Y_k+x))
\end{equation}

\subsection{Classification}\label{classifiers}

\begin{figure}[h]
    \centering
    \includegraphics[width=0.8\textwidth]{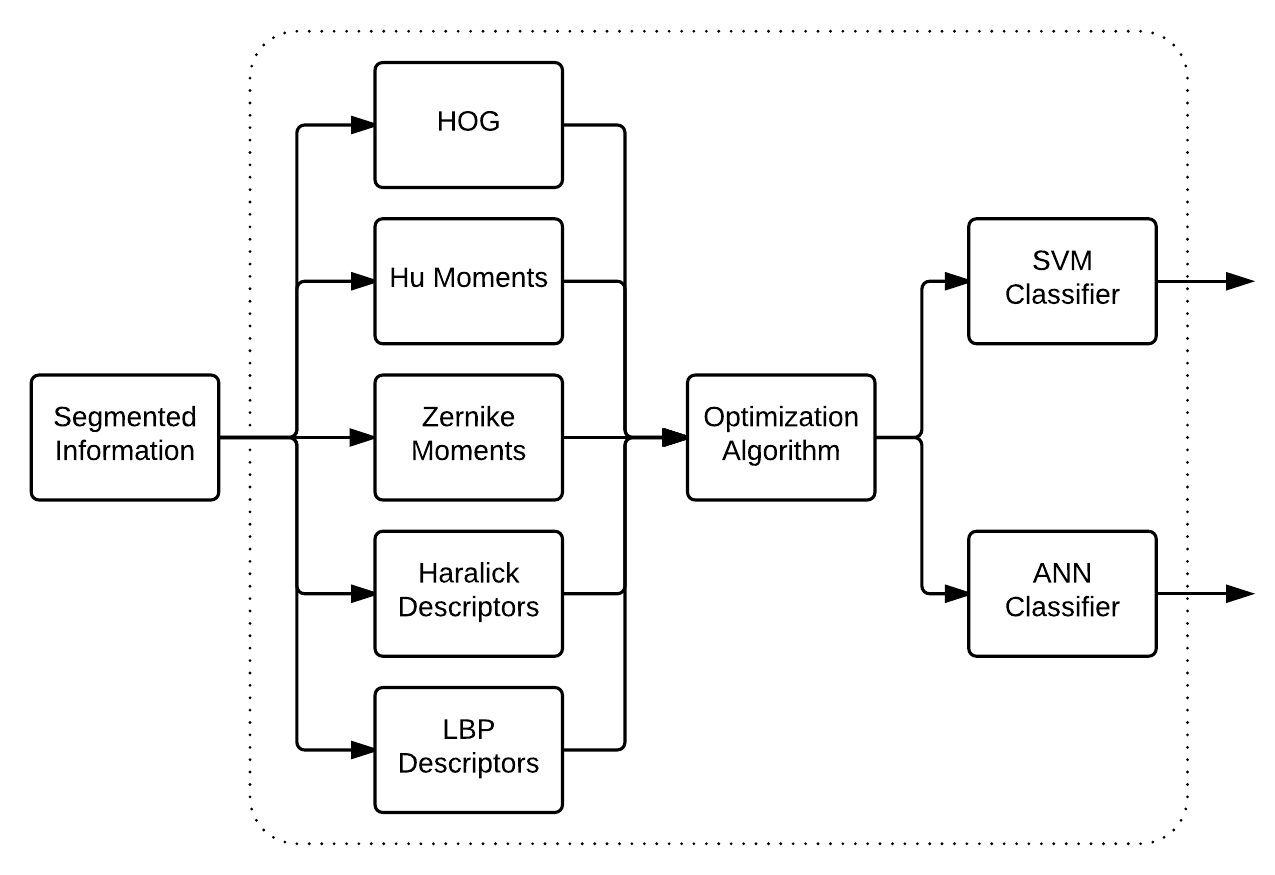}
    \caption{Algae classification methodology.}
    \label{fig:classDia}
\end{figure}

In this section the classification methods are explained. Classification is often the final step in pattern recognition. Methods such as Artificial Neural Networks (ANN) and Support Vector Machines (SVM) divide the space into a certain number of classes. Figure \ref{fig:classDia} shows the algae classification block, where the features extracted pass trough a classification stage. The Optimization Algorithm block is a feature mixture that uses SFS algorithm, this block is used in the experiments 6 and 12, and will be explained in the Experimental Framework section in detail.

\subsubsection{Artificial Neural Networks and Support Vector Machines}

ANN and SVM were used as classification techniques. The ANN used in this work have two hidden layers, $\tau$ is the number of neurons in each hidden layer and was defined as $\tau = 5,10,15,...,60$ trying to reach the best performance in the success rate of the neural network, the hidden layer structures are defined by $\sigma$, where $\sigma=\tau_1\tau_1,\tau_2\tau_2,...,\tau_{12}\tau_{12}$. A soft margin support vector machine (SVM) with Linear kernel was implemented too, where the complexity parameter ($C$) of the SVM and the bandwidth ($\gamma$) on its kernel are optimized in a exhaustive search up to powers of ten, with  $10^{-2}  \leq  C \leq  10^{2}$  and   $10^{-2}  \leq  \gamma \leq  10^{2}$.

\section{Experimental framework}

\subsection{Dataset}

The database contains 130 microscopical Neubauer chamber images of microalgae captured with a digital microscope. Original images were processed and segmented to obtain 4201 images of coenobia composed of one cell, 18035 images of coenobia composed of two cells, 19737 images of coenobia composed of four cells and 422 images of coenobia composed of eight cells. From the original database, 1680 images were extracted, 420 of each one (1-, 2-, 4- and 8-coenobium) which will be named as new database from now. This new database was extracted to avoid the unbalanced problem of the original database. The 1680 mentioned images were manually segmented to compare the performance of automatic and manual segmentation. Worth mentioning that all images of the new database are ensuring to have the same orientation. The database created is publicly available for download on the project link \cite{Salazar2015microalgaeDatabase}.

\subsection{Manual segmentation}

In computer vision the ground truth (GT) plays an important role in the evaluation process. The GT is important to develop new algorithms, to compare different algorithms, and to evaluate performance, accuracy and reliability \cite{fernandez2014semi}. For instance, in this paper the GT is the aforementioned manual segmentation images. To obtain the ground truth, simply an expert draws the contour of each algae on the original image. Therefore there are two versions of the new database, one with automatically segmented images with active contours procedure, and another with manually segmented images named ground truth images.

\subsection{Evaluation Metrics}\label{sec:metrics}

To evaluate the automatic segmentation performance, the Hoover metrics \cite{hoover1996experimental} were calculated. Hoover metrics consider five types of regions in the ground truth and machine segmented image comparison, either classified as correctly detected, over-segmented, under-segmented, missed and noise, and then plots the number of areas in each class weighted by total amount of areas based on a threshold (tolerance \%) term that is the free term in which the graphics are based. Other classical metrics are the precision, recall and f-measure, those metrics are originally applied to machine learning. The classical metrics were used on this paper to evaluate the segmentation performance like the Hoover metrics. Recall for example is the proportion of the real positive cases that are correctly predicted. Conversely, precision denotes the proportion of predicted positive cases that are correctly real positives. F-measure is the harmonic mean of recall and precision, the metric f-measure gives an idea of the accuracy of the test \citep{powers2011evaluation}.

To validate the classification performance, a cross-validation with ten folds was executed. The cross-validation consists of random separation of the features database on $k$ folds, train with $k-1$ folds, and validate with one fold changing this last one on each iteration, and finally calculate the mean and standard deviation of the $k$ experiments.

\subsection{Implementation details}

The Sequential Forward Selection algorithm and the Haralick, HOG, Hu and LBP descriptors were extracted using Balu Toolbox \cite{Mery2011Balu}. The Zernike moments were extracted with codes realized by Tahmasbi et al. \cite{tahmasbi2011classification,saki2013fast}. The multiclass SVM classifier was implemented with the LIBSVM library \cite{chang2011libsvm}.

\subsection{Experiments}\label{sec:experiments}

The experiments were carried out using the 5 features described in Section \ref{featuresExtraction} in an individual and combined way. For each segmented microalgae image, 7 Hu moments, 81 HOG, 40 Zernike moments, 59 LBP descriptors, and 28 Haralick features (mean and range) were extracted. A total of 420 algae of each class were considered in this work.

To validate each experiment, the aforementioned cross-validation was executed with $k=10$. Table \ref{tab:experiments} lists the experiments performed, where the 5 features explained, and the optimization algorithm SFS was tested with ANN and SVM classifiers, to find which one fits more to the problem.

\begin{table}[h]
\centering
\setlength{\tabcolsep}{6pt}
\caption{Description of the experiments.}
\label{tab:experiments}
\begin{tabular}{llllll}
\hline
Test & Features  & Classifier &Test & Features  & Classifier  \\
\hline
Exp1 &  Zernike   & ANN & Exp7  & Zernike  & SVM\\
Exp2 &  HOG       & ANN & Exp8  & HOG      & SVM\\
Exp3 &  Hu        & ANN & Exp9  & Hu       & SVM\\
Exp4 &  LBP       & ANN & Exp10 & LBP      & SVM\\
Exp5 &  Haralick  & ANN & Exp11 & Haralick & SVM\\
Exp6 &  All SFS   & ANN & Exp12 & All SFS  & SVM\\
\hline
\end{tabular}
\end{table}

In order to measure classification results the accuracy was computed from the confusion matrix. Each experiment listed in Table \ref{tab:experiments} was carried out 10 times of the cross-validation, and final results contain the mean and standard deviation of this repetitions with the best parameters (neurons on the hidden layer for ANN, $\gamma$ and $C$ for SVM, and the best $l$ features obtained from the SFS optimization technique for the experiments 6 and 12).

\subsubsection{Experiment 6, and experiment 12}

The experiments 6 and 12 were carried out as an exhaustive search of the $n$ features selected with the SFS algorithm, where $n=1,2,3,...,215$, the total amount of features. With the Equation \ref{eqn:argOptimization} the best $l$ features are found it, $M$ is the error of the classifier chosen, and $\vec{Y}_n$ are the $n$ best features chosen by the SFS algorithm. $\vec{E}$ is the matrix of the error and $\vec{S}$ is the matrix of the standard deviations. Finally the error and the standard deviation of the experiment 6 and 12 are chosen with the Equations \ref{eqn:error} and \ref{eqn:std}.

\begin{eqnarray}
\label{eqn:argOptimization}
l &=& \argmax_{n} M(\vec{Y}_n)\\
\label{eqn:error}
error &=& \vec{E}(l)\\
\label{eqn:std}
std &=& \vec{S}(l)
\end{eqnarray}

\section{Results and discussion}

In this section the segmentation and classification results are discussed. These results are exposed taking   the metrics explained in the Section \ref{sec:metrics} into account.

\begin{figure}[htbp]
    \centering
    \begin{subfigure}[b]{0.4\textwidth}
		\includegraphics[width=\textwidth]{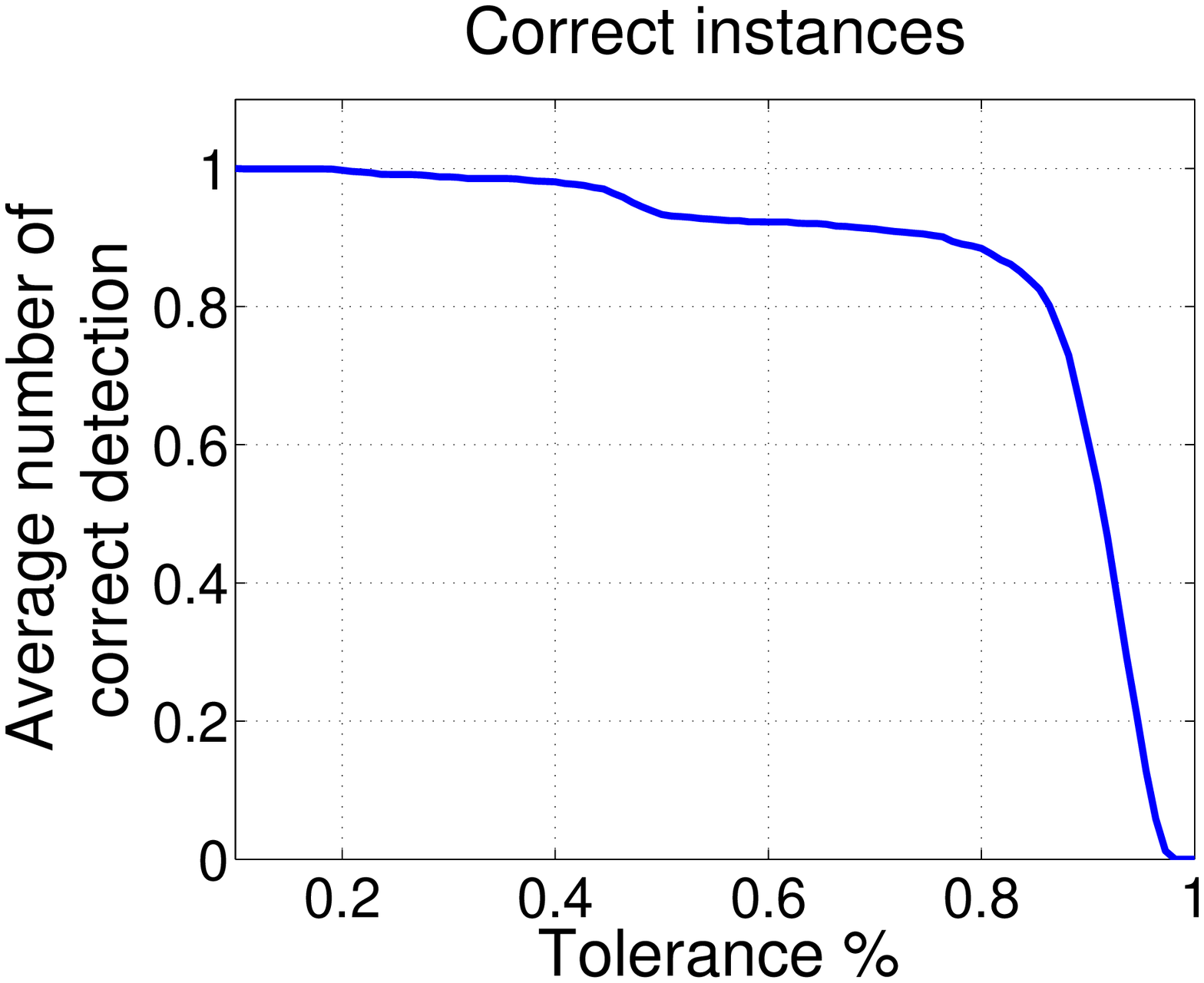}
        \caption{•}
        \label{fig:corrDetec}
    \end{subfigure}
    \begin{subfigure}[b]{0.4\textwidth}
		\includegraphics[width=\textwidth]{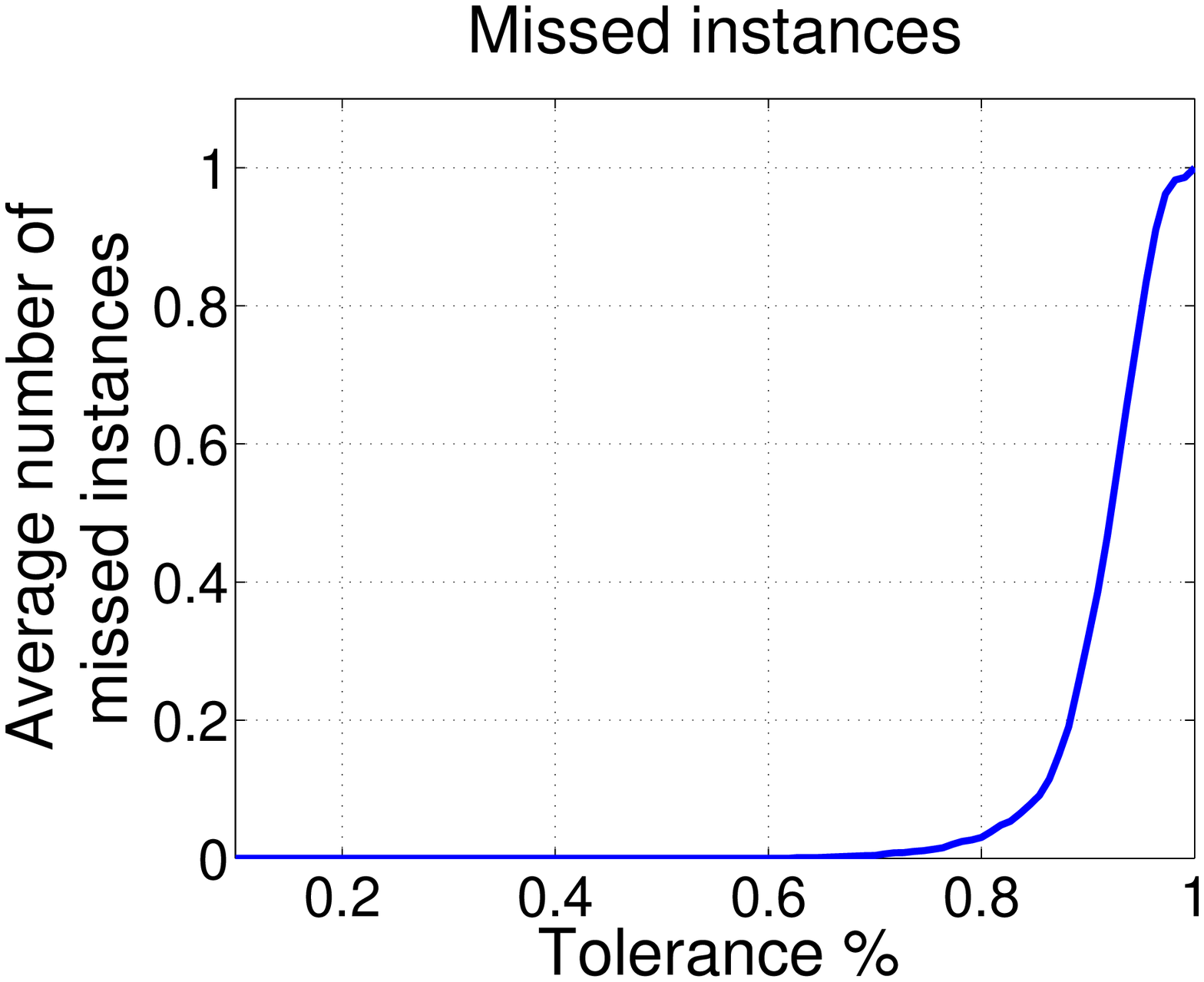} 
        \caption{•}
        \label{fig:missInst}
    \end{subfigure}
    \begin{subfigure}[b]{0.4\textwidth}
		\includegraphics[width=\textwidth]{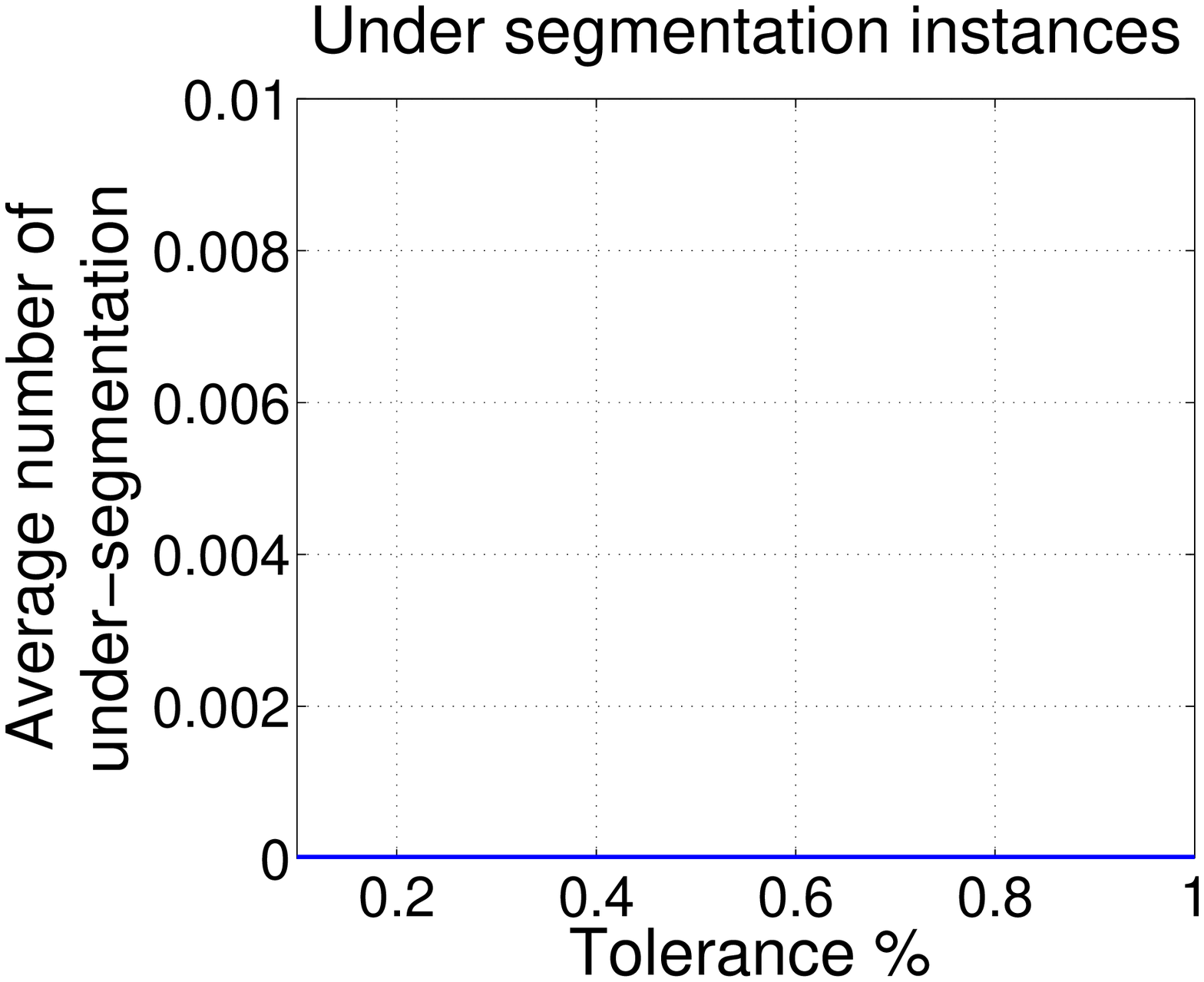} 
        \caption{•}
        \label{fig:underSegme}
    \end{subfigure}
    \begin{subfigure}[b]{0.4\textwidth}
		\includegraphics[width=\textwidth]{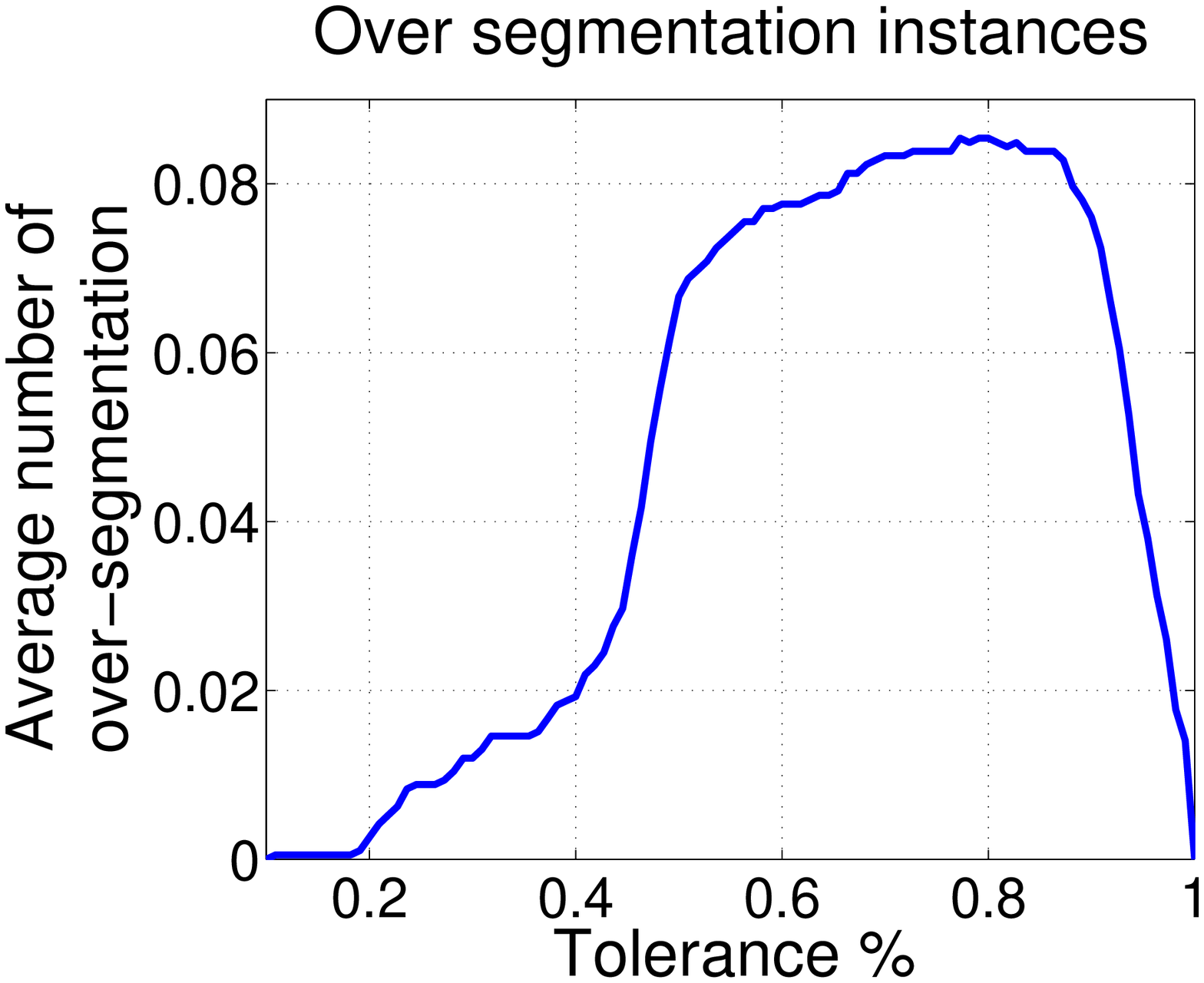} 
        \caption{•}
        \label{fig:overSegme}
    \end{subfigure}
    \begin{subfigure}[b]{0.4\textwidth}
        \includegraphics[width=\textwidth]{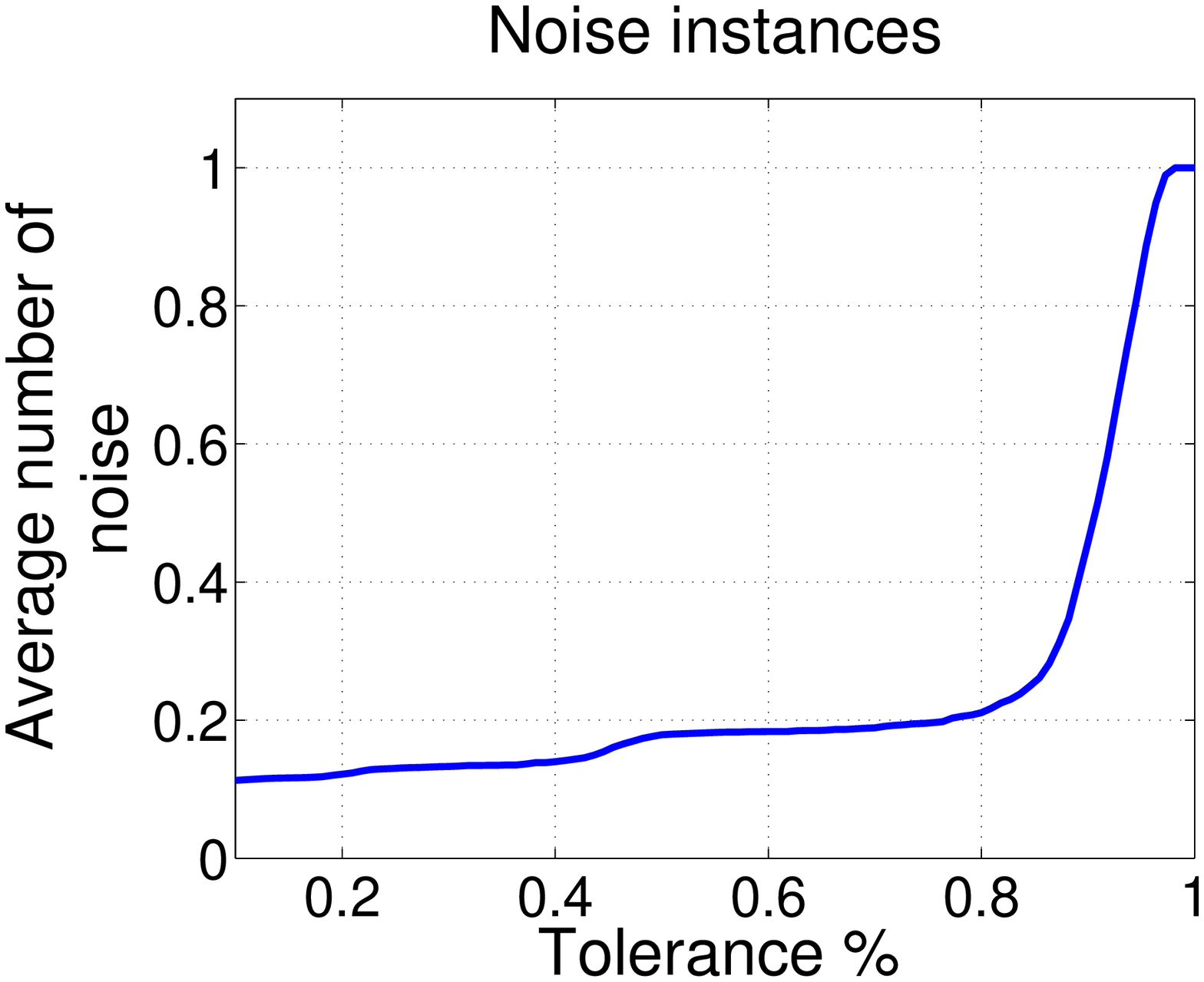}
        \caption{•}
        \label{fig:noise}
    \end{subfigure}
    \caption{Average of Hoover metrics.}
    \label{fig:HooverMetrics}
\end{figure}

Figure \ref{fig:HooverMetrics} shows the average Hoover metrics of the entire database. Each figure was extracted with the automatically segmented and ground truth images, and then the mean of each metric was extracted.

Figure \ref{fig:corrDetec} shows the average number of correct instances under a percentage of tolerance. The tolerance is the percentage of valid region necessary to categorize the segmentation in some instance. Figure \ref{fig:corrDetec} shows 88.44\% of average performance using active contours with 80\% of tolerance, i.e. with 80\% of tolerance, 88.44\% of the images were correctly segmented. There is a 0\% performance with 100\% of tolerance, due to the fact that there are no segmented images that are totally equal to some ground truth images. A 100\% of performance with 100\% of tolerance is an ideal, and is practically impossible due to border problem on the image format, or the difficulty to determine the borders exactly. Figure \ref{fig:missInst} shows 3.02\% of missed instances with 80\% of tolerance. For a tolerance less than 80\% there are no  missed instances or regions.


Figure \ref{fig:underSegme} demonstrates that there are no under-segmentation problems caused by the active contour approach. Figure \ref{fig:overSegme} shows that the over-segmentation problems are minimal. Figure \ref{fig:noise} shows less than 20\% of noise regions on average.

Table \ref{tbl:clasMetrics} shows the classical metric precision, recall and f-measure. Each metric was extracted with the database images, and the mean and standard deviation are exposed.

\begin{table}[h]
\centering
\caption{Classic metrics.}
\label{tbl:clasMetrics}
\begin{tabular}{cccc}
\hline
Metric & Precision          & Recall             & F-measure          \\ \hline
Value  & 92.53\% $\pm$ 5.33\% & 95.41\% $\pm$ 5.81\% & 93.71\% $\pm$ 3.50\% \\ \hline
\end{tabular}
\end{table}

The metric in Table \ref{tbl:clasMetrics}, commonly used for classification, shows that 95.41\% of the foreground is correctly predicted. The general performance of the automatic segmentation procedure is 93.71\% given by the f-measure. There is an unavoidable human error on the ground truth due to difficulties of segmenting the borders, and the small error margin in this microscopic images. The results in Figure \ref{fig:HooverMetrics} and Table \ref{tbl:clasMetrics} show that the active contour approach is correctly chosen. The results in Tables \ref{tbl:resultClass1} and \ref{tbl:resultClass2} validate the active contour segmentation approach, because the results of the Active Contour and the Ground Truth experiments are similar, i.e. the classification results using manual and automatic segmentation are close enough.


Tables \ref{tbl:resultClass1} and \ref{tbl:resultClass2} show the performance results of the experiments proposed in Table \ref{tab:experiments}. The proposed experiments 6 and 12 with the optimization algorithm show the best performances, that is a combination of the entire feature set extracted. It is important to mention that the feature set was standardized with z-score.

\begin{table}[h]
\centering
\setlength{\tabcolsep}{6pt}
\caption{Performance of the classification experiments with ANN.}
\label{tbl:resultClass1}
\begin{tabular}{lcccc}
\hline
\multirow{2}{*}{Test} & \multicolumn{2}{c}{Manual Aligned} & \multicolumn{2}{c}{Automatic Aligned} \\ \cline{2-5} 
                      & Active Contour & Ground Truth & Active Contour & Ground Truth \\ \hline
Exp1                  & 77.38 $\pm$ 2.49 \% & 76.01 $\pm$ 3.98 \% & 77.62 $\pm$ 2.85 \% & 76.49 $\pm$ 4.40 \% \\ 
Exp2                  & 83.57 $\pm$ 2.81 \% & 83.51 $\pm$ 2.16 \% & 83.81 $\pm$ 3.95 \% & 84.40 $\pm$ 2.28 \% \\ 
Exp3                  & 78.93 $\pm$ 3.52 \% & 80.36 $\pm$ 6.34 \% & 76.85 $\pm$ 7.31 \% & 82.32 $\pm$ 5.09 \% \\ 
Exp4                  & 89.46 $\pm$ 2.63 \% & 90.77 $\pm$ 3.14 \% & 92.02 $\pm$ 2.37 \% & 92.02 $\pm$ 1.44 \% \\ 
Exp5                  & \cellcolor[HTML]{FE0000}67.26 $\pm$ 4.26 \% & 73.81 $\pm$ 3.17 \% & 68.75 $\pm$ 3.10 \% & 75.42 $\pm$ 3.41 \% \\ 
Exp6                  & 97.14 $\pm$ 1.15 \% & 97.14 $\pm$ 0.83 \% & 97.20 $\pm$ 0.84 \% & \cellcolor[HTML]{32CB00}97.32 $\pm$ 1.20 \% \\ \hline
\end{tabular}
\end{table}

In Tables \ref{tbl:resultClass1} and \ref{tbl:resultClass2}, Automatic Aligned  refers to the experiments realized with the database aligned with the combination of the Fourier Transform and the Linear Regression.  Manual Aligned  refers to the experiments realized with the database with manually vertical aligned contours.  Active Contour refers to the experiments realized with the machine segmented images, and Ground Truth  refers to the experiments realized with the manually segmented images.

\begin{table}[h]
\centering
\setlength{\tabcolsep}{6pt}
\caption{Performance of the classification experiments with SVM.}
\label{tbl:resultClass2}
\begin{tabular}{lcccc}
\hline
\multirow{2}{*}{Test} & \multicolumn{2}{c}{Manual Aligned} & \multicolumn{2}{c}{Automatic Aligned} \\ \cline{2-5} 
                      & Active Contour & Ground Truth & Active Contour & Ground Truth \\ \hline
Exp7                  & 81.85 $\pm$ 2.21 \% & 79.64 $\pm$ 2.76 \% & 82.32 $\pm$ 3.49 \% & 79.52 $\pm$ 1.93 \% \\ 
Exp8                  & 89.17 $\pm$ 3.17 \% & 89.52 $\pm$ 2.30 \% & 90.24 $\pm$ 2.82 \% & 90.42 $\pm$ 2.74 \% \\ 
Exp9                  & 85.60 $\pm$ 2.69 \% & 88.69 $\pm$ 2.19 \% & 85.42 $\pm$ 2.93 \% & 88.21 $\pm$ 2.11 \% \\ 
Exp10                 & 93.10 $\pm$ 1.73 \% & 93.04 $\pm$ 1.59 \% & 93.87 $\pm$ 1.35 \% & 93.51 $\pm$ 0.91 \% \\ 
Exp11                 & \cellcolor[HTML]{FE0000}73.21 $\pm$ 2.71 \% & 79.17 $\pm$ 2.85 \% & 74.88 $\pm$ 3.35 \% & 78.75 $\pm$ 3.71 \%   \\ 
Exp12                 & 98.07 $\pm$ 0.89 \% & \cellcolor[HTML]{32CB00}98.63 $\pm$ 0.40 \% & 98.15 $\pm$ 0.52 \% & 98.21 $\pm$ 0.49 \% \\ \hline
\end{tabular}
\end{table}

It is important to see the difference between the ANN and SVM results in Tables \ref{tbl:resultClass1} and \ref{tbl:resultClass2}. The best and worst result using the ANN classifier was 97.32\% and 67.26\%, respectively. The best and worst result using the SVM classifier was 98.63\% and 73.21\%, respectively. Those results are a comparison among machine learning and statistical learning approaches. The statistical learning approach (SVM) shows better results than the machine learning approach. This means that the SVM found the optimal linear hyperplanes that separate the processed data.

Tables \ref{tbl:resultClass1} and \ref{tbl:resultClass2} show that results between manual aligned and automatic aligned are close enough to say that the manual alignment procedure is not necessary. Other words automatic approach and the Fourier transform is enough, nevertheless it is not a confirmation that descriptors are rotation independent.

The best performance is given by experiment 12, that is a mixture of all descriptors selected with the SFS algorithm.  In experiment 12 the result of the Equation \ref{eqn:argOptimization} is $l=65$. On the best 65 features, there are 17 LBP descriptors, 9 Haralick features, 3 Hu moments, 21 Zernike moments, and 15 HOG, this means 28.81\% of the entire LBP descriptors, 32.14\% of the all Haralick features, 42.86\% of the total of Hu moments, 52.5\% of entire Zernike moments, and 18.52\% of all HOG. Those results, with the 98.63\% of performance say that the features extracted were correctly chosen.

Table \ref{tbl:confusMat} shows the mean confusion matrix of the 10 folds cross-validation for the performance of 98.63\%. The problems occur between consecutive numbers of coenobia, e.g. the 0.48\% of coenobium of 1 cell is classified as coenobium of 2 cells, but there is no confusion with the coenobium of 4 and 8 cells with respect to the coenobium of 1 cell. The same occurs with the coenobium of 2 cells, where the confusion is with the 1-coenobia and 4-coenobia.

\begin{table}[]
\centering
\caption{Confusion matrix.}
\label{tbl:confusMat}
\begin{tabular}{ccccc}
\hline
            & 1 Cell     & 2 Cells      & 4 Cells      & 8 Cells      \\ \hline
1 Cell & 99.29 $\pm$ 1.61\% & 0.48 $\pm$ 1.51 \%  & 0 $\pm$ 0 \%        & 0 $\pm$ 0 \%        \\ 
2 Cells & 0.71 $\pm$ 1.61 \% & 97.62 $\pm$ 1.59 \% & 1.90 $\pm$ 2.19 \%  & 0 $\pm$ 0 \%        \\ 
4 Cells & 0 $\pm$ 0 \%       & 1.90 $\pm$ 1.51 \%  & 97.86 $\pm$ 2.08 \% & 0.24 $\pm$ 0.75 \%  \\ 
8 Cells & 0 $\pm$ 0\%        & 0 $\pm$ 0 \%        & 0.24 $\pm$ 0.75 \%  & 99.76 $\pm$ 0.75 \% \\ \hline
\end{tabular}
\end{table}

The automatic classification methodology presented in this work shows a maximum mean identification time of 2.43 $\pm$ 0.15 seconds by microalgae on a personal computer with one Intel Core i7 and 8 GB of memory RAM. This makes  the application of this methodology feasible as an alternative for  manual counting. Nevertheless, it allows the possibility to implement the experiment 6 or 12 in detail with the best $l$ features selected by the Equation \ref{eqn:argOptimization} with the results showed in Tables \ref{tbl:resultClass1} and \ref{tbl:resultClass2} to reduce the identification time.

\section{Conclusions}

We introduced an algorithm for Scenedesmus microalgae classification. The segmentation algorithm is composed of a histogram equalization, color quantization, and active contours iterations. The active contours algorithm shows results close enough to the manual segmentation procedure. The feature extraction step consists of extraction of the Hu and Zernike moments with the HOG, Haralick and LBP descritors to the segmented and original images. The classification is realized with Artificial Neural Network and Support Vector Machines classifiers. The SVM shows better results with respect to the ANN approach. The confusion matrix does not expose problems discriminating between coenobium in the classification. We reach 98.63\% of performance with SVM, ground truth images, and a mixture of features with SFS. The database derived from this work is publicly available for download.

For future work it is important to improve the classification performance and to reduce the process time and the amount of stages. The unbalanced problem on the acquisition step should be addressed with data augmentation techniques.

\section*{References}

\bibliography{mybibfile}

\end{document}